\documentclass{article}

\usepackage{arxiv}
\usepackage[utf8]{inputenc} % allow utf-8 input
\usepackage[T1]{fontenc}    % use 8-bit T1 fonts
\usepackage{hyperref}       % hyperlinks
\usepackage{url}            % simple URL typesetting
\usepackage{booktabs}       % professional-quality tables
\usepackage{amsfonts}       % blackboard math symbols
\usepackage{nicefrac}       % compact symbols for 1/2, etc.
\usepackage{microtype}      % microtypography
\usepackage{lipsum}
\usepackage{graphicx}
\graphicspath{ {./images/} }

\title{Discriminant Analysis in Contrasting Dimensions for Polycystic Ovary Syndrome Prognostication}

\author{
 Abhishek M. Gupta \\
  Bachelor of Engineering - EXTC\\
  University of Mumbai\\
  Mumbai, MH, IN \\
  \texttt{abhishekgupta@sjcem.edu.in} \\
   \And
 Himanshu H. Soni \\
 Bachelor of Engineering - EXTC\\
  University of Mumbai\\
  Mumbai, MH, IN \\
  \texttt{himansh056@gmail.com} \\
   \And
 Raunak M. Joshi \\
  Mentor\\
  Master of Engineering\\
  Mumbai, MH, IN \\
  \texttt{raunakjoshi.m@gmail.com} \\
  \And
 Ronald Melwin Laban \\
  Assistant Professor - Dept of EXTC\\ 
  St. John College of Engineering and Management\\
  Palghar, MH, IN\\
  \texttt{ronaldlaban@gmail.com} \\
}

\begin{document}
\maketitle
\begin{abstract}
A lot of prognostication methodologies have been formulated for early detection of Polycystic Ovary Syndrome also known as PCOS using Machine Learning. PCOS is a binary classification problem. Dimensionality Reduction methods impact the performance of Machine Learning to a greater extent and using a Supervised Dimensionality Reduction method can give us a new edge to tackle this problem. In this paper we present Discriminant Analysis in different dimensions with Linear and Quadratic form for binary classification along with metrics. We were able to achieve good accuracy and less variation with Discriminant Analysis as compared to many commonly used classification algorithms with training accuracy reaching 97.37\% and testing accuracy of 95.92\% using Quadratic Discriminant Analysis. Paper also gives the analysis of data with visualizations for deeper understanding of problem.
\end{abstract}

% keywords can be removed

\section{Introduction}
The Polycystic Ovary Syndrome also known as PCOS \cite{allahbadia2011polycystic} is a symptom which causes disorder in women ranging from reproductive age to their later stages in life. The commonly associated disorder with PCOS is irregularity in menstrual cycles and skin diseases. The high blood pressure and high cholesterol abide to the PCOS symptoms making things worse. High blood sugar is also observed in the later stages of the symptom. The early detection of this major disorder is a necessary issue. Machine Learning can be leveraged to the advantage of early detection. The medical expenses cannot be borne by all the women. Some also feel insecurity for disclosing such a matter even to doctors. A functional prognostication system can be very helpful in such a situation. Many people have proved to be successful in using Machine Learning with the right data for this problem. The PCOS is a binary classification problem. It clearly focuses on the result of whether the woman is suffering from PCOS or not. There is no distinction in the choices. Many machine learning algorithms can be used for such a problem. The first algorithm taken into consideration is Logistic Regression \cite{cramer2002origins} being the binary classification problem. A lot of work has been done on Logistic Regression so far now. Very brief implementation between Logistic Regression and Bayesian Classifier has been performed \cite{6139331}. The results obtained were having a Bayesian Classifier which was able to beat the performance of Logistic Regression with almost 2\% improvement. Trying to give more in depth implementation with machine learning inclination, various algorithms like KNN, CART and SVM were used \cite{8929674}. A detailed result section gives the metrics for the best algorithm. The improvements simultaneously were even extended with the boosting algorithms \cite{schapire2003boosting}. Many different boosting algorithms were used in an implementation for PCOS where the CatBoost was able to outrun all the boosting algorithms \cite{gupta2022succinct}. This was a compendious comparison of all the advanced boosting algorithms like AdaBoost, LGBM, XGBoost and CatBoost.

All methods that have been implemented by now are purely prognostication oriented with machine learning. What we present in this paper is to use Discriminant Analysis \cite{ghojogh2019linear} for getting a deeper understanding of the problem. Discriminant Analysis is an extension of dimensionality reduction with classification. Dimensionality reduction has supervised as well as unsupervised learning approaches \cite{Maaten2009DimensionalityRA}. The most commonly used dimensionality reduction technique is Principal Component Analysis \cite{jolliffe2016principal}. It is an unsupervised learning algorithm. The most commonly used supervised dimensionality reduction algorithm on the other hand is Linear Discriminant Analysis \cite{Tharwat2017LinearDA}. It maximizes the separability between the classes. An increase in dimensions gives us the Quadratic Discriminant Analysis \cite{Srivastava2007BayesianQD}. It is a type of generative model.

There are some differences between Linear Discriminant Analysis and Quadratic Discriminant Analysis. The observable difference is that Linear Discriminant Analysis does not have class-specific covariance matrices, but one shared covariance matrix among the classes. The shared covariance matrix is basically the covariance of all the inputs given. The benefit of Discriminant Analysis is that for the purpose of classification it uses the canonical variables.

\section{Implementation}
\subsection{Linear Discriminant Analysis}
The classification aspect of the Linear Discriminant Analysis works on maximum separability. Working with higher dimensions is necessary to generate a better understanding of the data for the final outcome. The higher dimensions not necessarily can give better understanding of the data points. So in order to maintain the feature dimensions and not increase them more than required, Linear Discriminant Analysis minimizes the dimensions. It transforms the features to one axis which is called Linear Discriminant. The common working of the algorithm considers the maximum separability by maximizing the difference between the means and minimizing the variance. This minimizing variation is called scatter in LDA The covariance matrix is the underlying source of the LDA. It approaches the problem by using conditional probability density functions which fall under normal distribution mean parameters and covariance parameters. For this the Bayesian Optimization \cite{frazier2018tutorial} is used which gives the log of likelihood for the ratios with threshold to classify the right class.

\subsection{Quadratic Discriminant Analysis}
The QDA \cite{Srivastava2007BayesianQD} is a variant of LDA \cite{Tharwat2017LinearDA} which considers an covariance matrix that is individual. It is estimated for all the classes of observations. If the individual classes orchestrates distinct covariances, in such a scenario QDA can prove to be better as there is an involvement of the prior knowledge. The effective parameters increase for QDA as compared to LDA. The most common consideration for the binary classification based Quadratic Discriminant Analysis or Linear Discriminant Analysis is given by a multivariate Gaussian Distribution \cite{Ahrendt2005TheMG}. The formula is given as follows

\begin{equation}
P(x | y=k) = \frac{1}{(2\pi)^{d/2} |\Sigma_k|^{1/2}}\exp\left(-\frac{1}{2} (x-\mu_k)^t \Sigma_k^{-1}(x-\mu_k)\right)
\end{equation}

The discriminant function for the LDA in Quadratic can be represented as follows

\begin{equation}
    \delta_k(x) = - \frac{1}{2} \log |\Sigma_k| - \frac{1}{2}(x- \mu_k)^T \Sigma_k^{-1} (x - \mu_k) + \log \pi_k\label{eq:1}
\end{equation}

We try to estimate $\Sigma_k$ in \ref{eq:1} for each and every class instead of assuming like we do in Linear Discriminant Analysis.

\section{Results}
This section constitutes our work with data. The implementation phase has many facets of results. We have presented numerous results below.

\subsection{Analysis}
This section of the paper clearly focuses on the analysis of the entire data. Many graphical representations given below provide an idea of different features in the data. The analysis section was taken into consideration to see the interaction of the different features with respect to age. The age is the parameter that influences the entire other parameters. The entire data can be checked for correlation using heat map. It gives a distinct distribution of the dataset features with valuation.

\begin{figure}[hbt!]
    \centering
    \includegraphics[scale=0.32]{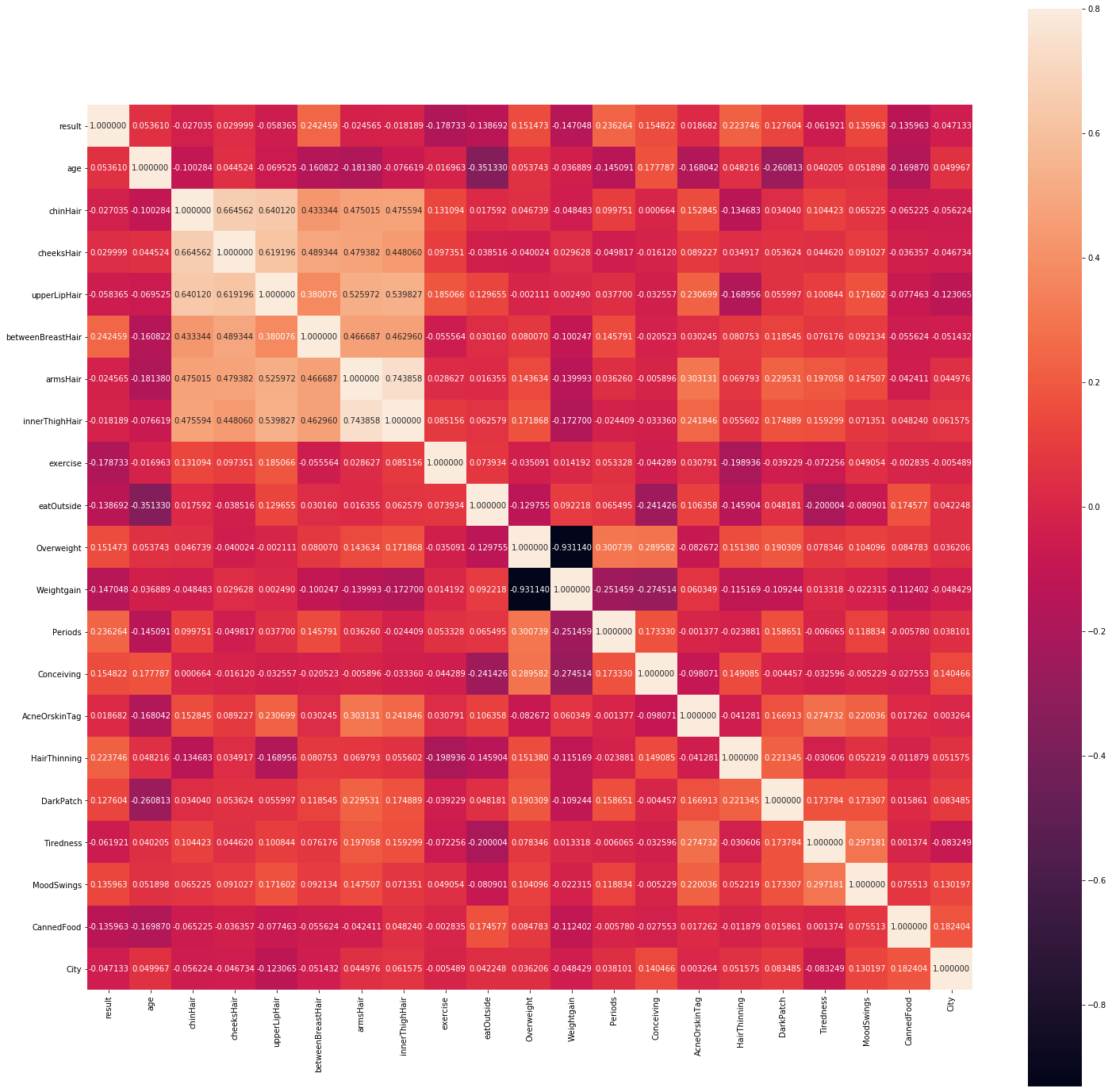}
    \caption{Heatmap for all the features}\label{fig:d}
\end{figure}

The Figure \ref{fig:d} is a very larger overview of all the features. It gives you an idea of how all the features are correlated from each other. More improvements can be bought into the analysis section. We can perform analysis for intricate observations. Boxplot \cite{Williamson1989TheBP} is known to be one of the most common form of visual representation to give a better edge in analysis of data.

\begin{figure}[hbt!]
    \centering
    \includegraphics[scale=0.7]{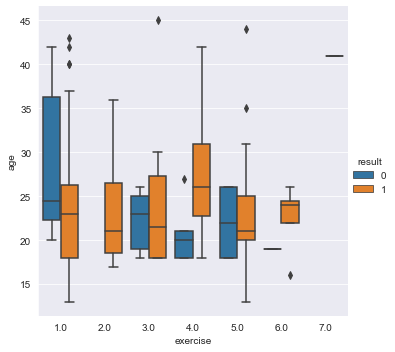}
    \caption{Categorical Plot for Exercise Level with respect to Age for Labels }\label{fig:e}
\end{figure}

\begin{figure}[hbt!]
    \centering
    \includegraphics[scale=0.7]{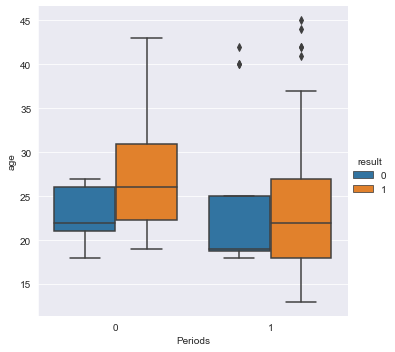}
    \caption{Categorical Plot for Periods with respect to Age for Labels }\label{fig:f}
\end{figure}

The Figure \ref{fig:e} gives a very good analysis of the effect of exercise levels with respect to age. The labels are taken into consideration. The visual representation of the features gives inference that exercise is an influencing factor taken into consideration for the prevention of the symptom. Another very important consideration is the irregularities in the menstrual cycles. It gives a very deep understanding about the problem. The Figure \ref{fig:f} gives a very detailed analysis about the distribution of the occurrence and effect of irregularities in menstrual cycles.

\subsection{Train and Test Accuracy}

The accuracies give the basic analogy of any algorithm. It the most primary metric taken into consideration. It gives a baseline effect of how well the model performs with the data.

\begin{table}[hbt!]
 \caption{Comparison of LDA and QDA Training \& Testing Accuracy}\label{tab:a}
  \centering
  \begin{tabular}{lll}
    \toprule
    Algorithm     & Training Accuracy     & Testing Accuracy \\
    \midrule
    Linear Discriminant Analysis & 86.84 \%  & 81.63 \%     \\
    Quadratic Discriminant Analysis     & 97.37 \% & 95.92 \%  \\
    \bottomrule
  \end{tabular}
\end{table}

In Table \ref{tab:a} we can clearly see that the variation in the accuracy of Quadratic Discriminant Analysis is comparatively less than Linear Discriminant Analysis.

\subsection{Confusion Matrix}
The Type-I and Type-II Error are the errors which give clear distinction of interacting variables with the test data. The basic aspect of Confusion Matrix \cite{ting2017confusion} is observation of predicted class with target class. It gives the errors which are targeted for specific problems. The Type-II Error works with False Negatives which is a type of error we want to target. The importance of this error is that if the person does have symptoms of PCOS, the system should detect the symptom. The table \ref{tab:b} below gives a comparison of False Positives and False Negatives between LDA and QDA.

\begin{table}[hbt!]
 \caption{Confusion Matrix for LDA \& QDA}\label{tab:b}
  \centering
  \begin{tabular}{lll}
    \toprule
    Algorithm     & False Negatives     & False Positives \\
    \midrule
    Linear Discriminant Analysis & 6  & 3     \\
    Quadratic Discriminant Analysis     & 0 & 0  \\
    \bottomrule
  \end{tabular}
\end{table}

\subsection{Metric Scores}

The two major metrics we are focusing in this paper are the Average Precision Score \cite{Zhang2009} and Precision Recall Curve \cite{10.1007/978-3-642-40994-3_29}. The Precision and Recall are individually calculated. Precision is a measure of quality of predictions. The Precision is given by formula

\begin{equation}
Precision = {\frac {True Positives}{True Positives + False Positives}}
\end{equation}

and Recall is given by formula

\begin{equation}
Recall = {\frac {True Positives}{True Positives + False Negatives}}
\end{equation}

Recall is a measure of the quantity of correctly classified predictions. Since our problem works with binary classification problem \cite{Kumari2017MachineLA}, the values are distinguished {\em Yes/No} and do not have any threshold values.

Average Precision considers a Precision-Recall curve as the weighted mean of all the Precision achieved at each threshold, with the increase in recall from the previous threshold used as the weight. The formula given below gives a precise representation for the binary classification problem. If the data is available with thresholds the graph considers an integral spread over the points. The discrete data on  the other hand gives the summation spread over the points. The formula is given as

\begin{equation}
Average Precision = \sum^{n}_{k=1}(Recall_k - Recall_{k-1}).Precision_k
\end{equation}

\begin{figure}[hbt!]
    \centering
    \includegraphics[scale=0.7]{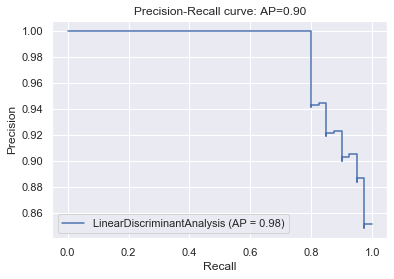}
    \caption{Precision Recall Curve with Average Precision for LDA}\label{fig:a}
\end{figure}

\begin{figure}[ht!]
    \centering
    \includegraphics[scale=0.7]{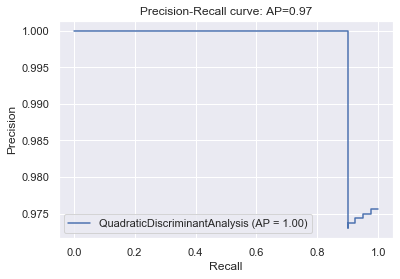}
    \caption{Precision Recall Curve with Average Precision for QDA}\label{fig:b}
\end{figure}

The below are given graphs that give clear distinction of above specified metrics in the form of visualization. These graphs plot the precision and recall score with respect to average precision. In Figure \ref{fig:a} we can see the interaction of the Precision and Recall Curve. The sudden drop can be seen after some points. Whereas in Figure \ref{fig:b} we can a sharp drop which goes beyond the performance of the Linear Discriminant Analysis.

\subsection{Receiver Operator Characteristic and Area Under Curve}
Receiver Operator Characteristic Curve \cite{FAWCETT2006861} also known as RoC Curve is used for performance measurement of classification models. It outputs comprehensive probability fluctuations throughout the predictions. This is computed on the basis of two functions known as True Positive Rate (TPR) and False Positive Rate (FPR). True Positive Rate is also known as Sensitivity and works much similar to Recall. It is denoted by formula as

\begin{equation}
True Positive Rate = {\frac {True Positives}{True Positives + False Negatives}}
\end{equation}

After True Positive Rate we have to work our way towards False Positive Rate, but the issue is that the calculation cannot be done directly. One needs to first calculate Specificity. It is a measure for finding negative classes without anomalies. This is represented by formula as

\begin{equation}
Specificity = {\frac {True Negatives}{True Negatives + False Positives}}
\end{equation}

After this specificity can help compute the False Positive Rate. It can be also denoted with a formula

\begin{equation}
False Positive Rate = {\frac {False Positives}{False Positives + True Negatives}}
\end{equation}

Apart from RoC Curve we require Area Under Curve which is also known as AUC \cite{BRADLEY19971145}. This AUC supports specifying threshold settings. It gives degree or measure of separability. Higher the value of AUC, better is the prediction. The higher the value on the scale, the more better algorithmic performance.

\begin{figure}[h]
    \centering
    \includegraphics[scale=0.625]{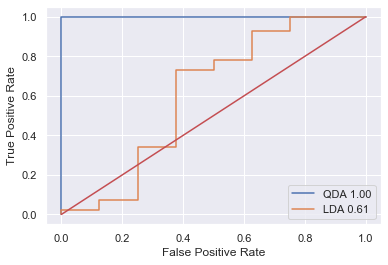}
    \caption{RoC Curve with AUC for LDA and QDA}\label{fig:c}
\end{figure}

In Figure \ref{fig:c} the Red Line indicates the boundary of threshold for the Area Under Curve. The metrics for the models need to be above the Red Line. In our paper, the Quadratic Discriminant Analysis performs better than Linear Discriminant Analysis. The QDA covers much more area than LDA and the inference can be derived that more area for classification is considered.

\subsection{Separability}
Considering Discriminant Analysis the representation of the points is given over plots with dimensions.

\begin{figure}[h]
    \centering
    \includegraphics[scale=0.6]{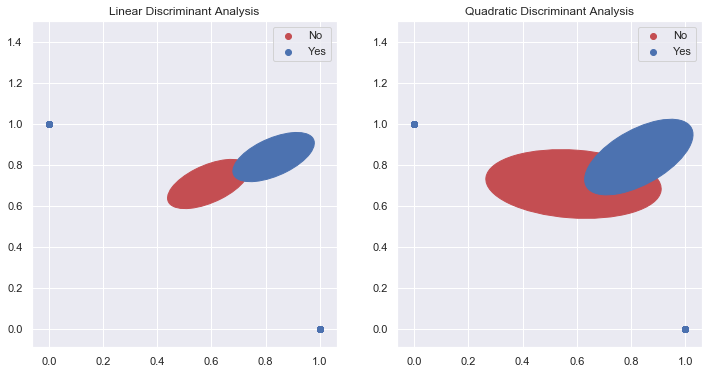}
    \caption{The Separability of the Data with respect to Labels}\label{fig:g}
\end{figure}

In Figure \ref{fig:g} we can see the separation in both the labels of their respective distributions. The blue ellipse represents the Yes labels and red ellipse represents the No labels. The Linear Discriminant Analysis has ellipses that are separated in Linear dimensional form while Quadratic Discriminant Analysis has ellipses in quadratic form, with much more increased dimensions.

\section{Conclusion}
This paper presents a very intricate area of Machine Learning domain which is not much worked with. Our aim in this paper was highlighting the point that classification with dimensionality reduction is as important as other classification parametric methods. We used supervised dimensionality reduction algorithm Linear Discriminant Analysis. We worked on increasing the dimension of the same algorithm which resulted in Quadratic Discriminant Analysis. Definitely this paper promotes more ideas and new improvements can be done by researchers in the future. With our best belief and knowledge we conclude this paper.

\bibliographystyle{unsrt}  
\bibliography{references}

\end{document}